\documentclass{article}

    \PassOptionsToPackage{numbers, compress}{natbib}

\usepackage[preprint]{neurips_2024}

\usepackage[utf8]{inputenc} 
\usepackage[T1]{fontenc}    
\usepackage[hidelinks]{hyperref}       
\usepackage{url}            
\usepackage{booktabs}       
\usepackage{amsfonts}       
\usepackage{nicefrac}       
\usepackage{microtype}      
\usepackage{xcolor}         
\usepackage{graphicx}
\usepackage{svg}

\title{Hybrid Student-Teacher Large Language Model Refinement for Cancer Toxicity Symptom Extraction}

\author{%
  Reza Khanmohammadi\thanks{Corresponding author: Reza Khanmohammadi (khanreza@msu.edu).} \\
  Michigan State University \\
  \texttt{khanreza@msu.edu} \\
  \And
  Ahmed I. Ghanem\\
  Henry Ford Cancer Institute \\
  \texttt{aghanem1@hfhs.org} \\
  \And
  Kyle Verdecchia \\
  Henry Ford Cancer Institute \\
  \texttt{kverdec1@hfhs.org} \\
  \And
  Ryan Hall \\
  Henry Ford Cancer Institute \\
  \texttt{rhall14@hfhs.org} \\
  \And
  Mohamed Elshaikh \\
  Henry Ford Cancer Institute \\
  \texttt{melshai1@hfhs.org} \\
  \And
  Benjamin Movsas \\
  Henry Ford Cancer Institute \\
  \texttt{bmovsas1@hfhs.org} \\
  \And
  Hassan Bagher-Ebadian \\
  Henry Ford Cancer Institute\\ 
  \texttt{hbagher1@hfhs.org} \\
  \And
  Bing Luo \\
  Henry Ford Cancer Institute \\
  \texttt{bluo1@hfhs.org} \\
  \And
  Indrin J. Chetty \\
  Cedars Sinai Medical Center\\
  \texttt{indrin.chetty@cshs.org} \\
  \And
  Tuka Alhanai \\
  New York University Abu Dhabi \\
  \texttt{tuka.alhanai@nyu.edu} \\
  \And
  Kundan Thind\thanks{Shared senior author.} \\
  Henry Ford Cancer Institute\\
  \texttt{kthind1@hfhs.org} \\
  \And
  Mohammad M. Ghassemi\footnotemark[2] \\
  Michigan State University \\
  \texttt{ghassem3@msu.edu} \\
}

\begin{document}

\maketitle

\begin{abstract}
Large Language Models (LLMs) offer significant potential for clinical symptom extraction, but their deployment in healthcare settings is constrained by privacy concerns, computational limitations, and operational costs. This study investigates the optimization of compact LLMs for cancer toxicity symptom extraction using a novel iterative refinement approach. We employ a student-teacher architecture, utilizing Zephyr-7b-beta and Phi3-mini-128 as student models and GPT-4o as the teacher, to dynamically select between prompt refinement, Retrieval-Augmented Generation (RAG), and fine-tuning strategies. Our experiments on 294 clinical notes covering 12 post-radiotherapy toxicity symptoms demonstrate the effectiveness of this approach. The RAG method proved most efficient, improving average accuracy scores from 0.32 to 0.73 for Zephyr-7b-beta and from 0.40 to 0.87 for Phi3-mini-128 during refinement. In the test set, both models showed an approximate 0.20 increase in accuracy across symptoms. Notably, this improvement was achieved at a cost 45 times lower than GPT-4o for Zephyr and 79 times lower for Phi-3. These results highlight the potential of iterative refinement techniques in enhancing the capabilities of compact LLMs for clinical applications, offering a balance between performance, cost-effectiveness, and privacy preservation in healthcare settings.
\end{abstract}

\section*{Introduction}
\subsection*{The Need for Optimized LLMs in Clinical Settings}
\noindent\textbf{Opportunities and Challenges:} The integration of Large Language Models (LLMs) into clinical informatics presents both significant opportunities and unique challenges \cite{MENG2024109713}. Clinical institutions face a critical need for advanced natural language processing capabilities, particularly in symptom extraction from unstructured texts. However, this need is tempered by several constraints inherent to the healthcare environment \cite{Ong}.

\noindent\textbf{Privacy Concerns and Resource Limitations:} Foremost among these constraints is the imperative to protect patient privacy \cite{Esmaeilzadeh}. This concern often leads clinical institutions to prefer on-premises deployment of LLMs, avoiding the risks associated with transmitting sensitive data to external servers \cite{Esmaeilzadeh}. However, this preference introduces its own set of challenges. Large-scale LLMs, while powerful, require substantial computational resources that many healthcare facilities find difficult to maintain and operate on-site \cite{Shah2023}. The associated costs can be prohibitive, both in terms of initial investment and ongoing operational expenses.

\noindent\textbf{Smaller LLMs and Their Limitations:} Consequently, there is a growing interest in smaller LLMs that are more suitable for local deployment \cite{fu2024tinytitanssmallerlarge}. These compact models offer a potential solution to the resource and cost constraints many clinical settings face. However, they come with their own limitations, primarily stemming from their reduced exposure to comprehensive clinical corpora \cite{Wu2024}. The relative scarcity of clinical text data and the typically smaller training sets in this domain exacerbate this challenge.

\noindent\textbf{Economic Barriers to Third-Party Solutions:} Furthermore, the use of third-party generative AI models through APIs, such as OpenAI's ChatGPT, while potentially powerful, often proves economically unfeasible for many clinical environments due to high usage costs \cite{Musser2023ACA}. This economic barrier emphasizes the need for efficient, locally deployable solutions.

\noindent\textbf{Optimization Challenges:} The primary challenge, therefore, lies in optimizing smaller LLMs to robustly process clinical data, despite constraints in computational capacity and training data availability. There is a pressing need to develop techniques that can enhance the performance of these compact models, particularly in tasks such as symptom extraction and analysis \cite{Pitfalls}. The goal is to achieve an optimal balance between preserving data privacy, managing operational costs, and ensuring high-fidelity performance in clinical applications.

\noindent\textbf{Demand for Novel Strategies:} This complex landscape of needs and constraints underscores the importance of developing innovative approaches to LLM refinement and deployment in clinical settings. It calls for solutions that can leverage the power of advanced language models while addressing the unique requirements and limitations of the healthcare environment.

\subsection*{Clinical Symptom Extraction with LLMs}
Recent studies have explored LLMs for extracting clinical information from unstructured EHR texts. Mahbub et al. \cite{Mahbub2024LeveragingLL} used zero-shot learning with Flan-T5 for SUD severity extraction, outperforming rule-based approaches. Reese et al. \cite{Reese2024} found GPT-4's performance in clinical diagnostics to be sensitive to prompt formulation. Shyr et al. \cite{Shyr2024} demonstrated ChatGPT's efficacy in zero- and few-shot settings for rare disease phenotype extraction. Guevara et al. \cite{Guevara2023LargeLM} showed fine-tuned Flan-T5 models' superiority in extracting social determinants of health, especially with synthetic data augmentation.
These studies underscore LLMs' potential to enhance critical health information extraction from clinical texts, improving symptom and phenotype identification for effective radiation oncology toxicity management.

\begin{figure*}[!t]
\centerline{\includegraphics[width=1.4\linewidth]{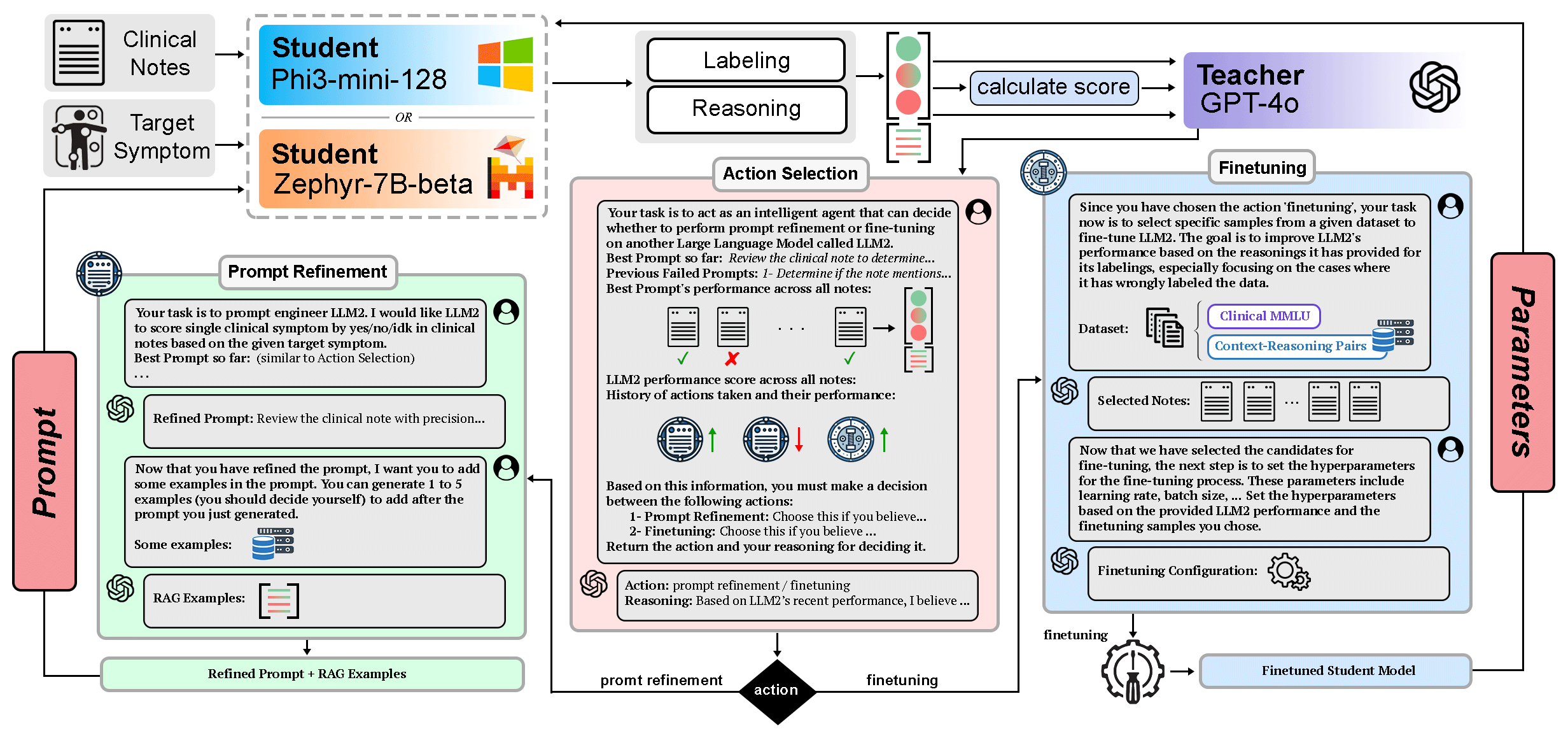}}
    \caption{The diagram illustrates the iterative refinement method, involving a student model (Phi3-mini-128 or Zephyr-7B-beta) and a teacher model (GPT-4o). The process starts with the student model receiving clinical notes and a target symptom, generating initial labels and reasoning. The teacher model then assesses performance and decides between prompt refinement and fine-tuning. In prompt refinement, the teacher improves the prompt and adds RAG examples. In fine-tuning, the teacher selects relevant samples and sets hyperparameters for the student model. In the hybrid approach, the teacher model acts as an intelligent agent, dynamically deciding between prompt refinement and fine-tuning based on the student's performance and needs. The refined approach is iteratively applied, optimizing the student model's performance in symptom extraction.}
\label{fig:hybrid-llm-refinement}
\end{figure*}

\subsection*{Iterative LLM Refinement Techniques}
\noindent\textbf{Single LLM Refinement:} Recent advancements in LLMs have focused on iterative refinement techniques to enhance performance across diverse tasks. Chen et al. \cite{Chen2023IterativeTR} demonstrated that multiple refinement rounds significantly improve translation fluency and naturalness. Building on this concept, Xiong et al. \cite{Xiong2024WatchES} developed the IPR framework, which outperformed established baselines in complex interactive tasks. Madaan et al. \cite{madaan2024self} further advanced this approach with Self-Refine, yielding significant improvements without additional training data. Yan et al. \cite{Yan2023RefiningTR} extended these principles to enable resource-efficient performance enhancements in less capable models. In the domain of bioinformatics, Chen et al. \cite{Chen2023IterativePR} applied iterative prompt refinement to substantially improve ChatGPT's accuracy in extracting gene relationships and biological pathways.

\noindent\textbf{Collaborative LLM Refinement:} Recent advancements have explored collaborative frameworks utilizing multiple LLMs. Lee et al. \cite{Lee2024LLM2LLMBL} introduced LLM2LLM, where a teacher LLM augments small datasets to enhance student LLM performance in low-data scenarios. Zhang et al. \cite{Zhang2024TSAlignAT} developed TS-Align, aligning LLMs with human preferences without manual annotations. Saha et al. \cite{Saha2023CanLM} demonstrated significant improvements in student LLM performance through teacher-generated personalized explanations. Yuan et al. \cite{Yuan2024BoostingSC} showed that teacher-generated analogies substantially improve student LLMs' scientific concept comprehension and question-answering capabilities. These studies collectively demonstrate the efficacy of collaborative LLM frameworks in enhancing model performance across diverse domains and tasks, particularly in scenarios with limited data or complex reasoning requirements.

\subsection*{LLM Refinement in Symptom Extraction}
Recent research has focused on integrating advanced LLM techniques with clinical symptom extraction, presenting significant potential for clinical informatics advancements. Clinical institutions prefer local LLM deployment for data privacy, but face challenges with the high costs and computational demands of large models. Smaller LLMs, while more suitable for on-premises use, are limited by reduced exposure to comprehensive clinical corpora. The primary challenge lies in optimizing these compact LLMs for robust clinical data processing, given constraints in computational capacity and data availability. Iterative refinement techniques and data augmentation strategies offer promising solutions to enhance LLM performance in clinical settings, aiming to balance data privacy, operational costs, and extraction accuracy. A notable attempt to address these challenges is presented in the work by Khanmohammadi et al. \cite{khanmohammadi2024iterativepromptrefinementradiation}, who introduce a novel student-teacher architecture, using Mixtral as the student model and GPT-4 as the teacher for prostate cancer radiotherapy symptom extraction. Their approach demonstrates significant improvements, highlighting the potential of advanced prompt engineering in LLMs for radiation oncology applications. 

\subsection*{Remaining Gaps}
Despite significant progress in integrating LLMs with clinical symptom extraction, several research gaps persist:
\begin{itemize}
\item The application of iterative student-teacher frameworks to smaller LLMs for on-premises clinical deployment remains unexplored.
\item Incorporation of Retrieval-Augmented Generation (RAG) \cite{ke2024developmenttestingretrievalaugmented} into prompt refinement techniques to enhance contextual understanding and accuracy.
\item Further investigation of fine-tuning for domain-specific adaptations within the student-teacher paradigm.
\item Expansion of the teacher's role in guiding the student model's fine-tuning process.
\end{itemize}
Addressing these gaps could provide valuable insights into the practical implementation and effectiveness of advanced LLM techniques in clinical settings, contributing to the broader discourse on healthcare informatics.
\subsection*{Contributions of This Work}
This study investigates several aspects of the student-teacher framework for clinical symptom extraction:
\begin{itemize}
\item Application of the framework to smaller LLMs (7 and 3.8 billion parameters) for clinical settings.
\item Integration of RAG within prompt refinement to enhance contextual understanding and accuracy.
\item Exploration of iterative fine-tuning for domain-specific adaptations in clinical contexts.
\item Implementation of an advanced student-teacher framework where the teacher model acts as a decision-making agent for optimizing refinement strategies.
\end{itemize}
These investigations aim to enhance the student-teacher framework for clinical symptom extraction, addressing current research gaps. The project implementation, including full prompt templates, is available on GitHub (removed for blind revision).
\section*{Experiments}
\subsection*{Data Description}
The dataset used in this study comprises clinical notes from prostate cancer patients definitively treated with 78 Gy radiotherapy (RT) between 2013 and 2020. We extracted 294 clinical notes documented beyond 6 months post-RT to focus on long-term toxicities, selecting notes that exhibited single symptoms. Our analysis concentrated on twelve common late post-RT toxicity symptoms: Cystitis, Dysuria, Erectile Dysfunction, Hematuria, Incontinence, Nocturia, Proctitis, Rectal Bleeding, Stricture, Urgency, Urinary Obstruction, and Urothelial Carcinoma. The dataset was structured with a training set of 20 notes per symptom (15 for Urothelial Carcinoma due to limited availability) and a test set of 5 notes per symptom (4 for Urothelial Carcinoma). Each note was labeled with a symptom assignment of -1 (absence), 0 (unknown status), or 1 (presence). The use of clinical notes and the overall study design were reviewed and approved by the institutional review board, ensuring ethical compliance and data protection standards were met throughout the research process.

In addition to the clinical notes, we specifically utilized selected categories from the Massive Multitask Language Understanding (MMLU) dataset \cite{mmlu} for fine-tuning. MMLU is a benchmark designed to assess models on a wide range of subjects. Our study focused on the following categories to form a clinical subset: anatomy, clinical knowledge, college medicine, human sexuality, medical genetics, and professional medicine. This selection resulted in a total of 1225 records, denoted as \(M\). We selected this data based on the Superficial Alignment Hypothesis \cite{NEURIPS2023_ac662d74}, which posits that a model’s knowledge and capabilities are largely learned during pretraining, while fine-tuning can effectively teach the model-specific subdistributions and formats using a relatively small set of examples. By applying this hypothesis to the clinical domain, we aim to enhance the model's familiarity and focus on clinical knowledge, thereby improving its performance in clinical symptom extraction tasks.
\subsection*{Data Preprocessing}
Our data preprocessing involves two key steps: embedding training clinical notes and generating context-reasoning (\(C_R\)) pairs. We utilize \texttt{Bio\_ClinicalBERT} to embed each clinical note in the training set into a 768-dimensional vector space, capturing the semantic content of the notes. These embeddings are then stored in a vector database for efficient retrieval during the refinement process.
To enrich the dataset, we employ GPT-4o to generate (\(C_R\)) pairs for each note. For every clinical note, its associated toxicity, and ground truth label, GPT-4o extracts the relevant textual context supporting the label and generates a reasoning explaining the labeling decision. These pairs are subsequently added as metadata to their corresponding note entries in the vector database.
This preprocessing approach serves to create a semantically rich dataset that combines dense vector representations with interpretable explanations, thus providing a robust foundation for our iterative refinement process and enabling more context-aware improvements to the student model's performance.

\subsection*{Iterative Refinement in the Student-Teacher Framework}

\subsubsection*{Concepts and Terminologies}
The iterative refinement process in a student-teacher framework involves two primary components:
\begin{itemize}
    \item \textbf{Student Model:} The student model, such as Phi3-mini-128 \cite{phi3} or Zephyr-7B-beta \cite{zephyr}, is responsible for the initial task of symptom extraction from clinical notes. The student model generates outputs based on specific inputs and prompts.
    \item \textbf{Teacher Model:} The teacher model, GPT-4o in our case, oversees the refinement process. It evaluates the performance of the student model, maintains a history of interactions, and generates refined prompts or fine-tuning configurations to improve the student model's performance iteratively.
\end{itemize}

\subsubsection*{Iterative Refinement Process}
As shown in Figure \ref{fig:hybrid-llm-refinement}, the iterative refinement process operates as follows:

\noindent \textbf{1. The student classifies notes} The student model receives a set of inputs: clinical notes, the target symptom (\texttt{\(S\)}), the prompt, and model parameters. The initial prompt template is structured as follows:
\begin{quote}
\textit{"Answer the following yes/no/idk question. Does the following clinical note mention the symptom of \(S\)?"}
\end{quote}
Using these inputs, the model processes each clinical note to extract and classify the target symptom as present (yes), negated (no), or unknown (idk). Crucially, the student model is also tasked with providing a reasoning for each classification, enhancing the interpretability of its outputs. This initial extraction serves as the baseline for subsequent refinement iterations.

\noindent \textbf{2. The student is assessed:} Student-generated labels are evaluated to calculate performance scores such as accuracy, precision, recall, and F1 score.
\begin{figure}[t]
    \centering  
    \centerline{\includegraphics[width=0.9\textwidth]{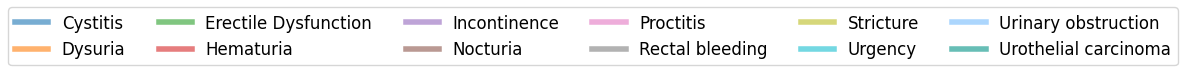}}
    \centerline{\includegraphics[width=0.9\textwidth]{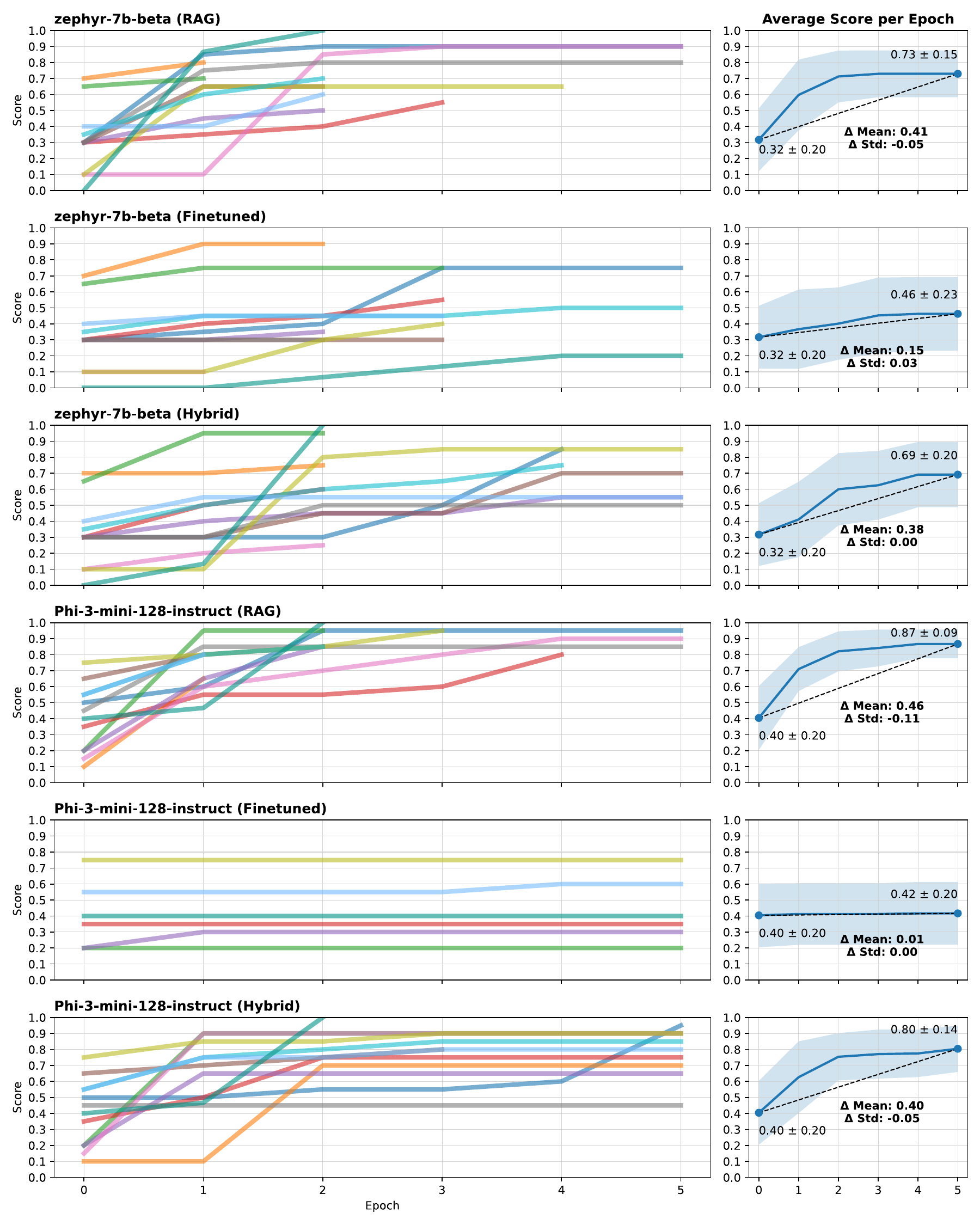}}
    \caption{Performance Comparison of Zephyr and Phi3 Models in Symptom Extraction. The line charts on the left-hand side represent the evolution of the Accuracy scores for different post-RT toxicity symptoms, with each line color-coded according to the 12 symptoms listed in the legend. The right-hand side line chart shows the average performance score across all toxicity symptoms at each time point, illustrating the difference in mean and standard deviation between the initial student model's performance and the final refined model's performance.}
    \label{fig:performance_comparison_train}
\end{figure}
\noindent \textbf{3. The teacher attempts to refine the student:} The performance scores, along with the generated prompts and reasoning, are sent to the teacher model. The teacher model reviews this information, considering the history of previous interactions and actions taken. Based on this analysis, the teacher model determines the most appropriate action to improve the student's performance.

\noindent \textbf{4. The teacher selects an action:} The teacher model selects one of the following actions:
\begin{itemize}
\item \textbf{Prompt Refinement:} If this action is chosen, the teacher model generates a refined prompt and may add RAG examples to enhance the contextual understanding of the student.
\item \textbf{Fine-Tuning:} If this action is chosen, the teacher model selects specific samples and sets hyperparameters for fine-tuning the student.
\end{itemize}

\noindent \textbf{5. The refined approach is applied and iterated:} The refined prompt or fine-tuning configuration is applied to the student model, and the next iteration begins. This iterative process continues, alternating between prompt refinement and fine-tuning (in the hybrid approach), with the goal of continually improving the student model's performance.

\noindent \textbf{6. Iterative Refinement:} The process is structured into epochs, each comprising 16 refinement rounds. In each round, the teacher model refines the prompt or fine-tunes the student model to improve symptom extraction. If performance improves, the epoch ends, and a new epoch begins with renewed refinement opportunities. The process terminates when no improvement is achieved across an entire epoch, indicating the student model's peak performance within the framework's constraints.

\subsection*{Methods Investigated}
\subsubsection*{Prompt Refinement Approach} \label{sec:prompt_refinement}
The prompt refinement approach is a key component of our iterative refinement process, focusing on optimizing the prompts used by the student models for symptom extraction. This approach consists of two main steps:

\noindent \textbf{1. Prompt Refinement:} When this action is selected, the teacher model (GPT-4o) is tasked with refining the prompt based on the student's performance. The teacher is given the following instruction:
\begin{quote}
\textit{"Your task is to prompt engineer another Large Language Model called LLM2. I would like LLM2 to score single clinical symptom by yes/no/idk in clinical notes based on the given target symptom. I am including the performance score of this extraction by LLM2 which is calculated based on ground truth labels and LLM2's output labels for symptom extraction across all notes. Your duty is to change the prompt such that LLM2's output gets improved by having more similar LLM2 output labels as the ground truth labels. More specifically, I will provide you with a Best Prompt and other samples that didn't work as well as the Best Prompt. [...]"}
\end{quote}
The teacher model analyzes the current performance and generates a refined prompt aimed at improving the student model's symptom extraction capability.

\noindent \textbf{2. RAG Example Generation:} Following prompt refinement, the process advances to RAG example creation. This phase begins by extracting semantically similar (\(C_R\)) pairs from a vector database. For each clinical note, we employ the same embedding model (\texttt{Bio\_ClinicalBERT}) to generate embeddings and query the database for the three most similar notes. The associated context and reasoning pairs from these neighbors are then presented to the teacher model. Utilizing this contextual information alongside the refined prompt, the teacher model generates one to five RAG examples to enhance the student model's understanding of the task and improve its performance in symptom extraction. 
\begin{quote}
"Now that you have refined the prompt, I want you to add some examples in the prompt. You can generate 1 to 5 examples (you should decide yourself) to add after the prompt you just generated. The generation of examples should be influenced by the model's performance and its connection with the refined prompt."
\end{quote}
Finally, the refined prompt and the generated RAG examples are concatenated, creating a comprehensive instruction set that combines refined task guidance with relevant, context-rich examples.

\subsubsection*{Fine-Tuning Approach} \label{sec:fine_tuning}
The fine-tuning approach aims to adapt the student model for enhanced performance in clinical symptom extraction. This process involves:

\noindent \textbf{1. Sample Selection:} The teacher model (GPT-4o) selects samples from both the clinical MMLU dataset \(M\) and the context-reasoning pairs \(C_R\) derived from our clinical notes. The selection is guided by the following prompt:
\begin{quote}
\textit{"Select specific samples to fine-tune LLM2, focusing on improving its performance in clinical symptom extraction. Choose samples that address LLM2's current weaknesses, especially cases where it has incorrectly labeled symptoms. Return a list of at least 10 indices from the provided datasets."}
\end{quote}
This step ensures the selection of relevant samples that target the model's shortcomings in symptom identification and classification.

\noindent \textbf{2. Fine-Tuning Configuration:} The teacher model then determines the optimal hyperparameters for fine-tuning:
\begin{quote}
\textit{"Provide hyperparameters for fine-tuning the student model on clinical symptom extraction. Consider the model's current performance and the selected samples. Return a JSON object with parameters including learning rate, batch size, number of epochs, and LoRA-specific settings."}
\end{quote}
Using the selected samples and specified hyperparameters (including learning rate, per-device train batch size, number of train epochs, gradient accumulation steps, LoRA-specific settings such as \textit{lora\_r}, \textit{lora\_alpha}, and \textit{lora\_dropout}, \textit{max grad norm}, \textit{weight decay}, \textit{learning rate scheduler type}, \textit{warmup ratio}, \textit{optimizer}, and \textit{target modules}), the student model undergoes fine-tuning. This process focuses on enhancing the model's ability to accurately extract and classify symptoms from clinical notes.

\begin{figure*}[t]
    \centering  
    \centerline{\includegraphics[width=1.4\textwidth]{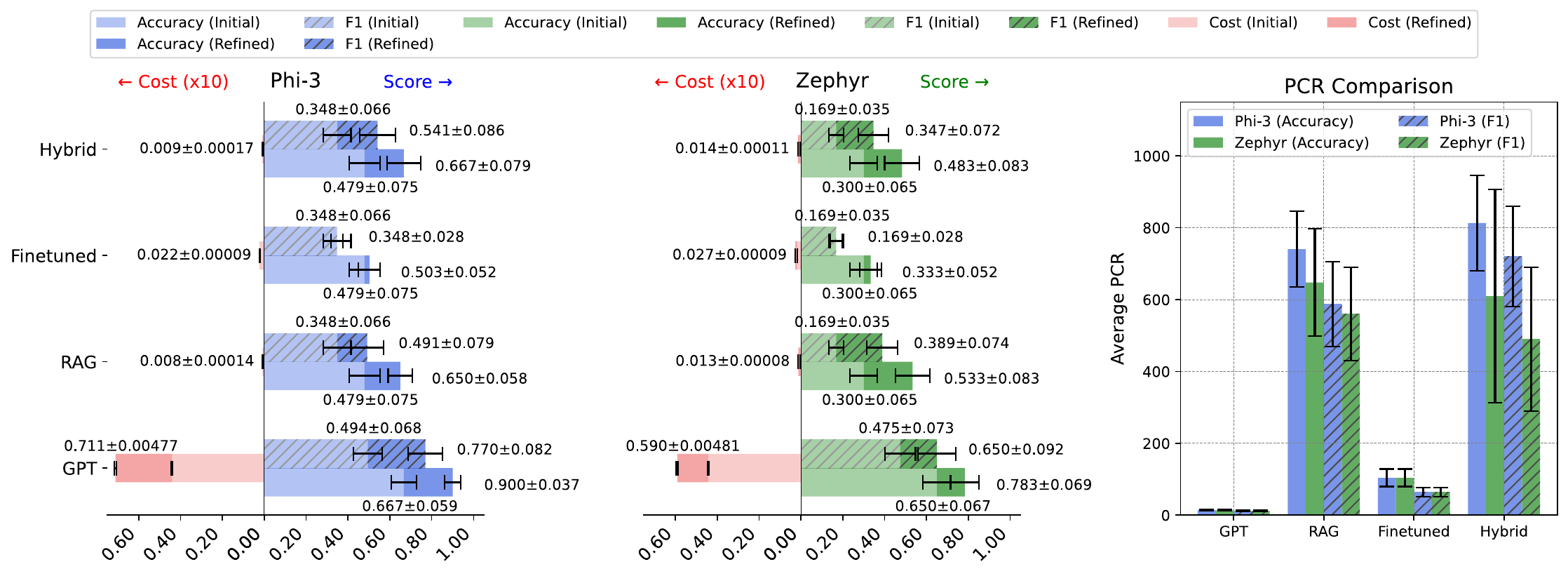}}
    \caption{The left panels display initial (brighter colors) and refined (darker colors) performance scores (blue and green bars) and associated costs (red bars) for the Phi-3 and Zephyr models across different refinement techniques: Hybrid, Finetuned, RAG, and GPT-4o. Hatched bars represent F1-macro scores, while smooth bars indicate accuracy. Averages and standard deviations are calculated across 12 toxicity symptoms. The right panel illustrates the average Performance-Cost Ratio for refined Phi-3 and Zephyr models, showing performance scores and associated costs.}
    \label{fig:performance_comparison_test}
\end{figure*}

\subsubsection*{Hybrid Approach}
The hybrid approach combines the strengths of both prompt refinement and fine-tuning, allowing for dynamic adaptation based on the model's current performance and needs. In this approach, the teacher model acts as an intelligent agent, deciding between prompt refinement and fine-tuning at each iteration. The process is as follows:

\noindent \textbf{1. Action Selection:} The teacher model is presented with a comprehensive prompt that includes:
\begin{itemize}
\item The best-performing prompt so far
\item Previous prompts that were less effective
\item The student model's performance across all notes, including ground truth labels, output labels, and reasoning
\item A history of previous actions taken and their resulting performance metrics
\end{itemize}
Based on this information, the teacher model is asked to choose between prompt refinement and fine-tuning. The decision prompt is structured as follows:
\begin{quote}
\textit{"Your task is to act as an intelligent agent that can decide whether to perform prompt refinement or fine-tuning on another Large Language Model called LLM2. LLM2 scores single clinical symptoms as yes/no/idk in clinical notes based on a given target symptom. [...] Based on this information, you must make a decision between the following actions: 
@prompt\_refinement: Choose this if [...] @finetuning: Choose this if [...]
Pay special attention to the reasonings LLM2 has provided for its labelings when making your decision."}
\end{quote}
The teacher model returns its decision as a JSON object, including the chosen action and a brief explanation for why it believes this action is the most effective next step.

\noindent \textbf{2. Action Execution:} Based on the teacher's decision:
\begin{itemize}
\item If prompt refinement is chosen, the process proceeds as described in Section \ref{sec:prompt_refinement}.
\item If fine-tuning is chosen, the process proceeds as described in Section \ref{sec:fine_tuning}.
\end{itemize}
This hybrid approach allows for a flexible and adaptive optimization strategy, leveraging the strengths of both prompt refinement and fine-tuning based on the specific needs and performance of the model at each stage of the process.
\subsection*{LLM Hyperparameter Selection for Text Generation}
The hyperparameters for the student and teacher models were selected to balance reproducibility in the student model with creative problem-solving in the teacher model. For the student model, we set the \textit{temperature} to 0.2, \textit{top-p} to 0.1, \textit{top-k} to 1, expected output length to 500 tokens, epochs to 5, and rounds per epoch to 16. These settings ensure deterministic outputs, minimizing randomness and enhancing reproducibility. For the teacher model, we set the \textit{temperature} to 1.9, \textit{top-p} to 0.9, \textit{top-k} to 50, and expected output length to 500 tokens. These values were chosen to allow for a diverse and exploratory generation of outputs, facilitating effective problem-solving and prompt refinement. These hyperparameters were chosen based on established guidelines for token selection and temperature settings in LLMs to optimize performance for our specific application.

\subsection*{Student and Teacher Evaluation Metrics}
In this study, we evaluate our student models by measuring accuracy and F1 macro scores after each symptom annotation in the training dataset. Additionally, the computational cost is assessed based on the prompt length and model weights. Costs are calculated using a function that determines the expenses of processing input and output tokens with GPT-4o, priced at \$5.00 and \$15.00 per million tokens respectively, reflecting the API's current rates at the time of writing. Energy consumption is also evaluated by tracking the power usage during model inference, translating this into kilowatt-hours and then into monetary costs, using the average U.S. electricity rate of 16.88 cents per kilowatt-hour. This methodical approach ensures a comprehensive understanding of both the clinical efficacy and operational efficiency of the deployed student models in realistic healthcare settings.

\section*{Results}
The performance evaluation of the Zephyr-7b-beta and Phi3-mini-128 models in extracting symptoms from clinical notes is presented across three refinement techniques: RAG, fine-tuning, and the hybrid method. These results, shown in Figure \ref{fig:performance_comparison_train}, highlight the improvement in symptom extraction accuracy over five epochs during the iterative refinement process of prompts, RAG examples, and model weights.

For the Zephyr-7b-beta model using the RAG method, the average score increased from 0.32 ± 0.20 to 0.73 ± 0.15, demonstrating a significant lift in performance with a mean increase of 0.41 and a standard deviation decrease of 0.05. Notable improvements were observed in symptoms like Urothelial Carcinoma, which increased from 0 to 1 in two epochs, and Proctitis, which rose from 0.1 to 0.9 in three epochs. In the fine-tuning approach, the average score improved to 0.46 ± 0.23, with a smaller increase in mean performance. The most improved symptom was Cystitis, which increased from 0.3 to approximately 0.8. The hybrid method achieved an average score of 0.69 ± 0.20, showing similar but slightly less improvement compared to RAG. Symptoms like Erectile Dysfunction showed more improvement in the hybrid method than in RAG, and Stricture improved from 0.1 to approximately 0.85, compared to 0.65 in RAG. 

For the Phi3-mini-128 model, the RAG approach resulted in the highest improvement, with the average score increasing from 0.40 ± 0.20 to 0.87 ± 0.09, showing a substantial lift of 0.46 in the mean and a decrease in standard deviation by 0.11. Fine-tuning showed limited improvements, with an average score of 0.42 ± 0.20, primarily in symptoms like Urinary Obstruction and Incontinence. The hybrid method for Phi3-mini-128 achieved a mean score of 0.80 ± 0.14, with a decrease in standard deviation by 0.05. Noteworthy improvements were seen in symptoms such as Nocturia, which increased to 0.9 in just one iteration, indicating faster convergence compared to the RAG method.

The results in Figure \ref{fig:performance_comparison_test} present the performance of Phi-3 across different refinement techniques on the test set for all toxicity symptoms, showing both accuracy and F1-macro scores along with associated average costs per test note. The initial accuracy for Phi-3 with the base prompt and weights was 0.48 across all methods. After refinement, the Hybrid approach showed the highest improvement with an accuracy of 0.67, closely followed by RAG at 0.65, while fine-tuning yielded minimal improvement to 0.50. For F1-macro scores, the initial performance was 0.35, with the Hybrid method achieving the highest refined score of 0.54, followed by RAG at 0.49, while fine-tuning showed no improvement. GPT-4o, tested with both the initial Phi-3 prompt and the refined Hybrid prompt (selected as it performed best in both accuracy and F1-macro), demonstrated initial accuracy and F1-macro scores of 0.67 and 0.49 respectively, improving to 0.90 and 0.77 with the refined prompt. Notably, the average costs per test note varied significantly: GPT-4o was the most expensive at \$$7.11 \times 10^{-2}$, followed by fine-tuning at \$$2.2 \times 10^{-3}$, Hybrid at \$$9.0 \times 10^{-4}$, and RAG being the most cost-effective at \$$8.0 \times 10^{-4}$.

The results for the Zephyr model, as shown in Figure \ref{fig:performance_comparison_test}, illustrate performance across various refinement techniques on the test set for all toxicity symptoms, presenting accuracy and F1-macro scores alongside average costs per test note. The initial accuracy for Zephyr with the base prompt and weights was 0.30 across all methods. Post-refinement, the RAG approach demonstrated the highest improvement with an accuracy of 0.53, followed by the Hybrid method at 0.48, while fine-tuning showed minimal improvement to 0.33. For F1-macro scores, the initial performance was 0.17, with RAG achieving the highest refined score of 0.39, followed by Hybrid at 0.35, and fine-tuning showing no substantial improvement at 0.17. GPT-4o, evaluated using both the initial Zephyr prompt and the refined RAG prompt (chosen for its superior performance in both accuracy and F1-macro), showed initial accuracy and F1-macro scores of 0.65 and 0.48 respectively, improving to 0.78 and 0.65 with the refined prompt. The average costs per test note varied considerably: GPT-4o was the most expensive at \$$5.90 \times 10^{-2}$, followed by fine-tuning at \$$2.7 \times 10^{-3}$, Hybrid at \$$1.4 \times 10^{-3}$, and RAG being the most cost-effective at \$$1.3 \times 10^{-3}$.

The right panel in Figure \ref{fig:performance_comparison_test} presents the average PCR for refined Phi-3 and Zephyr models across the 12 toxicity symptoms, with error bars indicating standard deviation across symptoms. For accuracy-based PCR, Phi-3 consistently outperformed Zephyr across all methods. The Hybrid approach yielded the highest PCR for Phi-3 at 813, followed closely by RAG at 740, while Zephyr achieved 647 for RAG and 609 for Hybrid. Fine-tuning showed identical modest PCR values of 104 for both models. GPT-4o, despite its high performance, had the lowest PCR (14 for Phi-3, 13 for Zephyr) due to its significantly higher cost. F1-based PCR values showed a slightly different pattern, with Phi-3 achieving 719 for Hybrid and 560 for RAG, while Zephyr scored 587 for RAG and 489 for Hybrid. Fine-tuning showed lower F1-based PCR at 65 for Phi-3 and 63 for Zephyr. These results underscore the efficiency of the Hybrid and RAG approaches in balancing performance and cost, with Phi-3 demonstrating superior cost-effectiveness across most methods, particularly in the Hybrid approach for F1-based metrics.


\section*{Conclusion and Discussion}
This study explored iterative refinement techniques to optimize compact, locally deployable LLMs for clinical settings, aiming to balance data privacy, computational constraints, and operational costs. Using a student-teacher architecture with Zephyr-7b-beta and Phi3-mini-128 as student models and GPT-4o as the teacher, we evaluated prompt refinement with RAG, fine-tuning, and a hybrid method.

Our findings revealed that the RAG method provided significant performance improvements and cost-efficiency. This approach substantially enhanced accuracy and F1-macro scores while maintaining low operational costs, making it a practical solution for clinical applications. Fine-tuning, on the other hand, was less effective when used alone, suggesting its role is best as a complementary technique rather than a standalone solution. The hybrid method, which combines RAG and fine-tuning, demonstrated the best overall performance in Phi3-mini-128 and outperformed other approaches in various scenarios. This method effectively harnessed the strengths of both RAG and fine-tuning, particularly for specific symptoms where combined strategies were essential.

Interestingly, our results showed that the smaller Phi3-mini-128, with approximately half the parameters of Zephyr-7b-beta, generally outperformed Zephyr in the task of clinical symptom extraction. This outcome suggests that model size alone does not determine effectiveness; rather, the refinement technique plays a crucial role in performance optimization.

The test results underscored the substantial cost implications associated with different models. Although GPT-4o delivered the highest absolute performance, its significantly higher expenses---45 times those of Zephyr and 79 times those of Phi-3---led to the lowest PCR. This finding underscores the importance of cost-effectiveness in practical applications, particularly in resource-constrained clinical environments.

In summary, our study demonstrates the effectiveness of iterative refinement techniques, particularly hybrid and RAG methods, in enhancing compact LLMs for clinical symptom extraction. These approaches balance performance and cost, offering viable alternatives to larger models. Our findings highlight the potential of advanced refinement techniques in improving LLM capabilities in data-sparse, specialized domains like healthcare. These methods effectively address performance needs and resource constraints, with potential applications beyond healthcare.
\bibliographystyle{IEEEtran}
\bibliography{references}
\end{document}